# DARTS: A Drone-Based AI-Powered Real-Time Traffic Incident Detection System

Bai Li[1], Achilleas Kourtellis[2], Rong Cao[3], Joseph Post[1], Brian Porter[4], Yu Zhang[1]*,

[1] Department of Civil and Environmental Engineering, College of Engineering, University of South Florida, Tampa, FL, 33620, USA

[2] Center for Urban Transportation Research, College of Engineering, University of South Florida, Tampa, FL, 33620, USA

[3] School of Civil and Environmental Engineering, Nanyang Technological University, 50 Nanyang Ave, 639798, Singapore

[4] District Seven, Florida Department of Transportation, Tampa, FL, 33612, USA

*Corresponding author: Address as above. Email address: yuzhang@usf.edu

E-mail addresses: baili@usf.edu (B. Li), kourtellis@usf.edu (A. Kourtellis), rong021@e.ntu.edu.sg (R. Cao), japost@usf.edu (J. Post), Brian.Porter@dot.state.fl.us (B. Porter), yuzhang@usf.edu (Y. Zhang).

## ABSTRACT

Rapid and reliable incident detection is critical for reducing crash-related fatalities, injuries, and congestion. However, conventional methods, such as closed-circuit television, dashcam footage, and sensor-based detection, separate detection from verification, suffer from limited flexibility, and require dense infrastructure or high penetration rates, restricting adaptability and scalability to shifting incident hotspots. To overcome these challenges, we developed DARTS, a drone-based, AI-powered real-time traffic incident detection system. DARTS integrates drones' high mobility and aerial perspective for adaptive surveillance, thermal imaging for better low-visibility performance and privacy protection, and a lightweight deep learning framework for real-time vehicle trajectory extraction and incident detection. The system achieved 99% detection accuracy on a self-collected dataset and supports simultaneous online visual verification, severity assessment, and incident-induced congestion propagation monitoring via a web-based interface. In a field test on Interstate 75 in Florida, DARTS detected and verified a rear-end collision 12 minutes earlier than the local transportation management center and monitored incident-induced congestion propagation, suggesting potential to support faster emergency response and enable proactive traffic control to reduce congestion and secondary crash risk. Crucially, DARTS's flexible deployment architecture reduces dependence on frequent physical patrols, indicating potential scalability and cost-effectiveness for use in remote areas and resource-constrained settings. This study presents a promising step toward a more flexible and integrated real-time traffic incident detection system, with significant implications for the operational efficiency and responsiveness of modern transportation management.

**INTRODUCTION**

Road traffic incidents not only result in fatalities and injuries but also lead to traffic congestion and secondary crashes, causing extensive transportation delay, excessive energy consumption and corresponding environmental pollution. According to the World Health Organization (WHO), road traffic accidents claim approximately 1.19 million lives globally each year and result in 20 to 50 million non-fatal injuries. The loss of productivity due to disabilities caused by non-fatal injuries further costs countries up to 3% of their annual GDP (*1*). Regarding the impact of incidents on road traffic, the Office of Operations of the Federal Highway Administration (FHWA) has identified traffic incidents as the No. 1 source of non-recurrent congestion (*2*). The additional vehicle emissions during congestion exacerbate urban air pollution, contributing to health problems and premature deaths among residents (*3*). In recent years, international organizations such as the World Bank and the United Nations have emphasized the importance of improving road traffic safety, particularly in developing and resource-constrained regions. For example, the World Bank's Global Road Safety Facility (GRSF) (*4*) and the UN Sustainable Development Goals (SDGs) (*5–8*) have highlighted road safety as a critical global concern. These efforts reflect a growing international recognition of the societal and economic impacts of traffic incidents, including their contribution to fatalities, injuries, and incident-induced congestion-related delays.

Prompt emergency care following a traffic crash is crucial, as even a few minutes of delay can mean the difference between life and death (*1, 9*). A study conducted across 2,268 counties in the United States found a significant association between longer emergency medical service (EMS) response times and higher motor vehicle crash mortality rates (*10*). Therefore, timely and accurate traffic incident detection and verification, combined with immediate access to on-site information for dispatching emergency responders, is essential for reducing the number of severe injuries and fatalities caused by traffic incidents (*11*). In terms of traffic flow restoration, if Transportation Management Centers (TMCs) can promptly obtain accurate information about the impact of incidents on upstream and downstream traffic flows, they can make timely decisions such as rerouting, adjusting traffic signals, or even closing sections of the road. These measures help mitigate the spread of non-recurrent congestion caused by traffic incidents, facilitate the rapid restoration of traffic flow, and reduce emissions and energy consumption associated with incident-induced congestion (*12, 13*). These demands underscore the critical importance of efficient and accurate traffic incident detection, which has gained increasing recognition in recent years for its vital role in saving lives and managing congestion.

Traditional incident detection methods can be categorized into microscopic and macroscopic approaches. Microscopic algorithms use trajectory data to detect traffic behaviors and incidents, while macroscopic methods analyze road traffic metrics such as flow, speed, occupancy, and kinetic energy and their variations to detect incidents (*14*). When an incident occurs, occupancy increases while volume and speed decrease upstream, whereas both occupancy and volume decrease downstream (*15*). These differences in upstream and downstream traffic features have been the basis of classic automated incident detection algorithms. In terms of traffic data collection for incident detection, inductive loop detectors (*16*) and floating vehicles with on-board GPS devices (*17, 18*) are the most commonly used methods. However, such methods rely heavily on the high-density deployment of sensors and vehicles to achieve accurate incident detection, even with the advancement of machine learning (ML) modeling (*19–22*). Additionally, the changes in traffic patterns and congestion caused by incidents often take time to propagate upstream, making it challenging to detect traffic incidents promptly and to assess their impact range accurately. Moreover, these sensing devices cannot provide an onsite scene view for verifying the incident and accurate evaluating the severity of the incidents.

With the widespread adoption of road traffic Closed-Circuit Television (CCTV) surveillance cameras and the rapid development of ML and deep learning (DL) models, particularly the You Only Look Once (YOLO) model (*23, 24*), some researchers have leveraged real-time video from surveillance cameras combined with ML/DL algorithms to analyze video data for automatic traffic incident detection, achieving a mean detection accuracy of approximately 90% (*25–35*). In the areas with no CCTV camera or CCTV cameras covering limited space, floating vehicles with dash cameras were used to collect traffic information. Some studies explored traffic incident detection using the videos from dash cameras in conjunction with ML/DL models, also achieving a mean detection accuracy of approximately 90% (*36–38*). Compared to loop detector and on-board GPS, cameras allow visual assessment of incident severity, however, accurate incident detection still rely on the high-density deployment of CCTV cameras with well refined observing angles or a sufficient level of penetration of floating vehicles with dash cameras. Furthermore, these detection methods lack an accurate and efficient approach to determining the impact range of an incident and the resulting propagation of congestion. On the other hand, visible-light RGB cameras used by CCTV systems and floating vehicles perform badly under low-visibility conditions such as nighttime and foggy weather (*31*). Additionally, visible-light RGB cameras may inadvertently capture sensitive information, such as license plates or driver faces, raising potential privacy concerns.

In light of these limitations, the advancement of drone/unmanned aerial vehicle (UAV) technology has provided an innovative platform for intelligent traffic monitoring and emergency response in smart cities (*39*). By analyzing aerial video data captured by drone-mounted cameras, researchers can efficiently extract both macroscopic and microscopic traffic data from urban networks, enabling flexible and efficient traffic sensing and parameter estimation (*40–48*). In the context of traffic incident-related drone research, Chen et al. utilized drones to hover over intersections or specific road segments to capture traffic monitoring videos (*49*). Based on these videos, they extracted vehicle trajectory distributions to observe safe spatial boundaries between vehicles and identified potential traffic conflicts when these boundaries were violated. Other studies have focused on high-precision on-site evidence collection and damage analysis for traffic incidents using drones (*50*, *51*). In summary, the use of drones for road traffic incident management is still in its early stage, and current research primarily emphasizes post-incident investigation and analysis, with particular attention to 3D reconstruction of accident scenes (*52*). However, there has been limited exploration of drone use for real-time traffic incident detection.

Based on the aforementioned literature review, these state-of-the-art traffic incident detection methods fail to achieve timely and accurate detection of traffic incidents, real-time online access to incident scene for severity assessment, and precise evaluation of the impact range and propagation of non-recurrent congestion caused by incidents. Given that traffic incidents can cause significant disruptions to traffic flow and markedly increase the likelihood of secondary incidents (*53*), these limitations underscore a critical gap in traffic incident detection.

To fill in the gap, this study developed a drone-based, AI-powered real-time traffic incident detection system. This system enables simultaneous real-time detection, online visual verification for severity assessment, and monitoring of congestion propagation caused by incidents. Specifically, our study accomplished the following: 1) Real-Time Traffic Incident Detection: To overcome low-visibility challenges and address privacy concerns, this study integrated drones equipped with thermal cameras. We developed an efficient incident detection framework to continuously extract vehicle trajectories from drone-captured thermal video streams and generate trajectory images at a fixed period. These images were then processed by a customized lightweight DL model to extract traffic flow features, enabling real-time detection of both traffic incidents and incident-induced non-recurrent congestions while accurately distinguishing them from recurrent congestions. 2) Online Incident Verification and Severity Assessment: Building on the incident detection framework, we developed the Drone-based AI-powered Real-time Traffic incident detection System (DARTS), which is an integrated software-hardware system featuring a web-based interactive Graphical User Interface (GUI). The GUI triggers an alarm when an incident is detected, automatically displaying the time of occurrence. This allows TMCs to promptly access the incident scene and its precise map location for real-time verification and severity assessment. The integrated process of detection, verification, and severity evaluation facilitates the timely dispatch of emergency responders, potentially reducing the risk of severe injuries and fatalities caused by delayed response. 3) Non-Recurrent Congestion Impact and Propagation Evaluation: DARTS is also capable of detecting the impact range and propagation speed of incident-induced non-recurrent congestion by calculating the length of the non-recurrent congestion sequence across different detection flights. Meanwhile, the GUI provides real-time visualization of congestion dynamics, thereby supporting TMCs in making prompt traffic management decisions to mitigate congestion propagation and reduce the likelihood of secondary incidents.

**METHODOLOGY**

To develop DARTS, we first designed a drone-based freeway traffic monitoring workflow to collect thermal traffic monitoring video data under incident, recurrent congestion and normal traffic conditions. Building on this foundation, we designed a traffic incident detection framework and a traffic incident detection platform for real-time detection of traffic incidents and visualization of detection results. Finally, we conducted field tests to evaluate DARTS's performance in detecting traffic incidents. The following sections provide a detailed description of the methodology proposed in this study.

**Research Architecture**

The research architecture of DARTS is illustrated in **Figure 1**. It is comprised of three main modules: data collection and preparation, the traffic incident detection framework, and the traffic incident detection platform. The data collection and preparation module is responsible for collecting high-quality thermal traffic monitoring videos on freeways under incident, recurrent congestion and normal traffic conditions. This is achieved by operating drones at optimal cruising altitude and speed, camera azimuth and pitch angles to cruise along freeways under varying traffic conditions. The thermal videos captured by the drones are then segmented into 2-minute clips, which are fed into the traffic incident detection framework for processing and incident detection.

The traffic incident detection framework processes video segments from the data collection and preparation module to detect traffic incidents. This framework consists of four components: (1) vehicle trajectory extraction and trajectory image generation; (2) incident detection using the proposed DL model; (3) aggregation of image-level detection results to generate a video-level detection result; and (4) extraction of incident features, including the incident-induced non-recurrent congestion length, its propagation speed, and the time period during which the incident scene is captured in the video segment. After developing and training the vehicle trajectory extraction and image generation algorithm, the traffic incident detection DL model, the image-to-video aggregation method, and the incident feature extraction algorithm, this framework is integrated into the traffic incident detection platform as a crucial component of the software system.

Finally, the traffic incident detection platform seamlessly integrates drone hardware, drone-side software, and workstation-side software systems. This unified platform supports real-time incident detection, automated incident feature extraction, and web-based visualization of detection results while providing the human–computer interaction function.

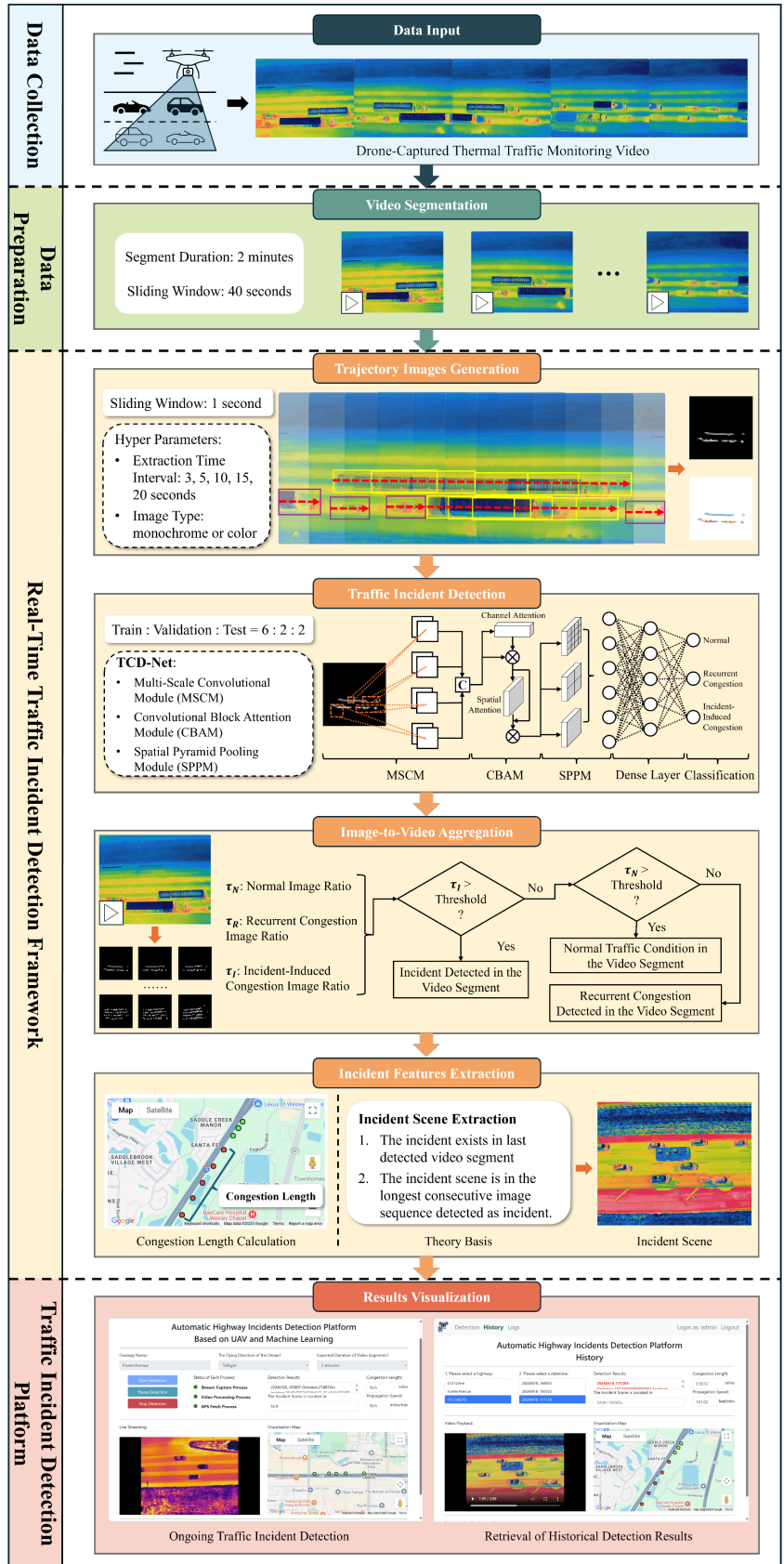

**Figure 1 Research architecture of DARTS.**

**Data Collection and Preparation**

High-quality thermal traffic monitoring videos are crucial for efficiently training algorithms and models within the traffic incident detection framework, thereby achieving optimal incident detection accuracy. In our preliminary studies, extensive experiments were conducted to evaluate the real-time vehicle detection performance of thermal videos captured by drones operating under various combinations of cruising altitudes, speeds, camera azimuths and pitch angles. Results indicated that a flight altitude of 200 feet with cruising speed of 10 miles per hour, a camera azimuth of 90° and a pitch angle between 70° and 90° yield the most balanced vehicle identification accuracy and stability (*54*). Therefore, during the collection of the thermal traffic monitoring video dataset and in the deployment of DARTS, these parameters were fixed to ensure standardized video collection, thereby optimizing incident detection accuracy.

To ensure that the framework accurately detects traffic incidents, it was imperative to collect thermal traffic monitoring videos from freeways under various conditions, including incidents, recurrent congestion, and normal conditions, for the training of algorithms and models. Accordingly, during the data collection process, we closely monitored freeway traffic conditions reported by the local TMC and promptly deployed drones to incident locations upon receiving incident reports to capture videos during incident conditions. In addition, videos were also collected under recurrent congestion and normal traffic conditions to form a comprehensive dataset.

Finally, in the practical deployment of DARTS, the thermal traffic monitoring stream is transmitted in real time. To ensure both real-time detection and discrete visualization of detection results, the data preparation module segments the continuous video stream into 2-minute video segments and inputs them into the traffic incident detection framework. Moreover, a 40-second sliding window is applied during segmentation to enhance the temporal and spatial resolutions of traffic incident detection.

**Traffic Incident Detection Framework**

The traffic incident detection framework, forming the core of DARTS, comprises four interdependent components: trajectory image generation, traffic incident detection, image-to-video aggregation, and incident feature extraction. Together, these components operate synergistically to enable real-time, automated detection of traffic incidents and extraction of incident features.

*Trajectory Image Generation*

The trajectory image generation component identifies vehicles within thermal video segments and extracts their movement trajectories. This process generates a set of vehicle trajectory images extracted at a fixed time period using a one-second sliding window, thereby providing a data foundation for subsequent DL model training. **Figure 2(a)** presents a flowchart outlining the complete process of vehicle identification, trajectory extraction, and image generation within each fixed time period, while **Figure 2(b)** illustrates the procedure and final output of a single trajectory image.

Specifically, due to the limited performance of pre-trained YOLO for thermal traffic monitoring videos, vehicles within the self-collected thermal video dataset were manually annotated to retrain the YOLO model and obtain adapted weight parameters (*55*). The custom-trained weights configuration was then applied to thermal video segments to extract vehicle coordinates frame-by-frame. Subsequently, the Lucas-Kanade optical flow tracker was employed to track the movement of each vehicle across consecutive frames (*56*), generating coordinate sequences that record vehicle movement. Because the drone used for video capture was also cruising along the freeway, motion compensation was performed to eliminate the drone's motion from the vehicle trajectory coordinate sequences. Then, vehicle movement directions were analyzed to exclude trajectories from opposing lanes, thereby avoiding the incorporation of disparate traffic features that could confound incident detection. Lastly, extracted vehicle trajectories within the time period were mapped onto a trajectory image, after which the algorithm proceeds to the next extraction period. To augment the sample size of the trajectory image generation component and enhance the accuracy of traffic incident detection, a one-second sliding window approach was adopted. For instance, if the extraction period is set to 20 seconds, the algorithm extracts trajectories for intervals $[t, t + 20], t = 0, 1s, 2s$ .... This sliding window method effectively increases the sample size and the robustness of DARTS by minimizing the adverse impact of isolated misclassifications of trajectory images on overall detection results.

Furthermore, two hyperparameters need to be optimized by experiments in this component: the vehicle trajectory extraction period and the image mode (monochrome or color) of generated trajectory images. The extraction period affects the completeness, smoothness, and coverage of trajectories, which in turn influence the DL model's ability to capture vehicle moving patterns and traffic flow features, thereby impacting the accuracy of incident detection. Regarding image mode, monochrome images emphasize structural and edge features, whereas color images highlight color information; each can affect feature extraction differently. To evaluate these effects and determine the optimal parameters, trajectory image datasets were generated using five different extraction periods (3 s, 5 s, 10 s, 15 s, and 20 s) and both monochrome and color image modes for model training and testing (**Figure 2**).

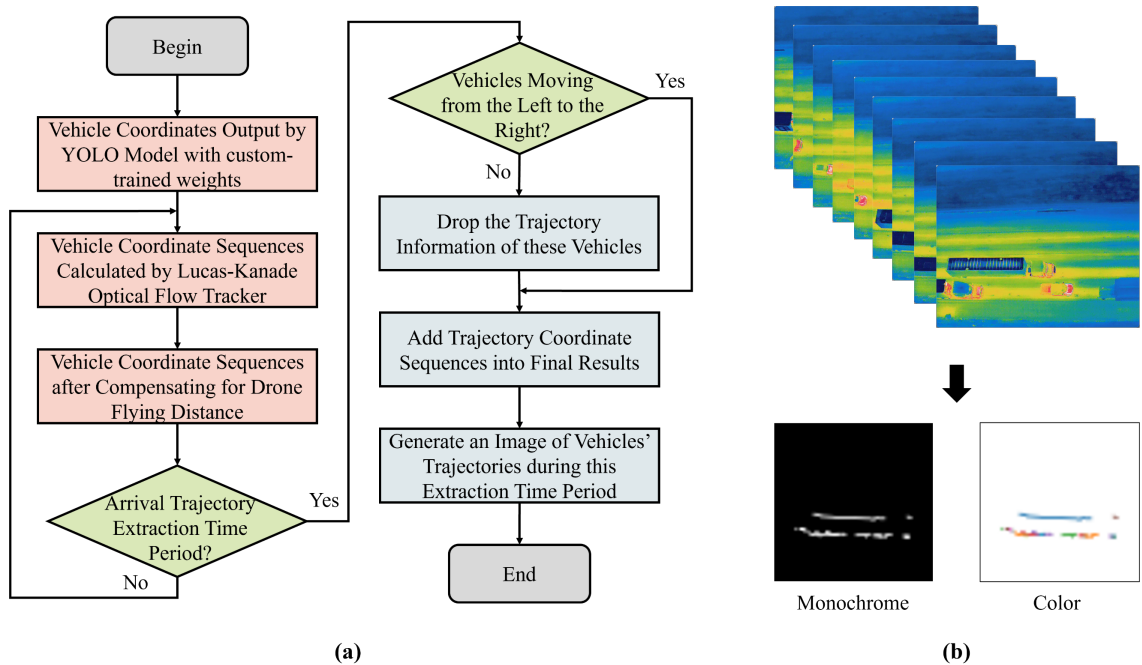

**Figure 2 Illustration of the trajectory image generation component. (a) Flowchart of trajectory image generation algorithm. (b) Visualization of the trajectory image generation.**

*Traffic Incident Detection Model*

Building on the trajectory image dataset, this study employs a DL model to automatically extract traffic features and achieve real-time detection of both traffic incidents and incident-induced non-recurrent congestions while accurately distinguishing them from recurrent congestions on freeways, an approach that deviates from traditional methods. Since the three traffic conditions (incident, recurrent congestion, and normal traffic) exhibit markedly different vehicle speeds, lane-changing behaviors, and vehicle density distributions, these differences are reflected in trajectory distributions on vehicle trajectory images. Moreover, deep neural networks can automatically extract these features to classify the traffic condition, effectively transforming the task into a three-class classification problem on extracted vehicle trajectory images. Considering the widespread application and superior performance of Convolutional Neural Networks (CNNs) in image recognition and classification (*57–59*), we designed a deep learning model based on the CNN architecture, augmented with multiscale CNN structure, Convolutional Block Attention Module (CBAM) (*60*) and Spatial Pyramid Pooling (SPP) (*61*), to perform this classification task—referred to as the Traffic Condition Detection Network (TCD-Net).

The architecture of TCD-Net is illustrated in **Figure 3**. The network primarily comprises a feature extraction module and a fully connected dense layers module. In the feature extraction module, a multiscale CNN is employed that utilizes four distinct convolutional kernels to efficiently extract features corresponding to short trajectories, long trajectories, and lane-changing trajectories from the trajectory image. To address the issues of redundant information and insufficient local sensitivity in CNN-based feature extraction, we integrate a CBAM module into the base CNN. This module sequentially applies channel and spatial attention mechanisms to adaptively recalibrate intermediate features, thereby emphasizing critical channel information and key local regions. Following three layers of multiscale CNN and CBAM feature extraction, we incorporate an SPP module to alleviate the loss of global context and inadequate capture of spatial hierarchies typically caused by single-scale pooling. The SPP module performs pooling at three different scales concurrently and concatenates the resulting features, ensuring that both global contextual and local detailed information are retained. Finally, the high-dimensional features extracted through multiple multiscale CNN and CBAM layers are transformed into a one-dimensional vector by the SPP module, and then reduced in dimensionality and nonlinearly mapped via fully connected dense layers. This process not only integrates and fuses the features but also reduces the model's complexity by decreasing the feature dimensions, and the resulting compressed features are ultimately mapped to the predefined class space through a SoftMax classifier, thereby achieving accurate trajectory image classification and traffic incident detection.

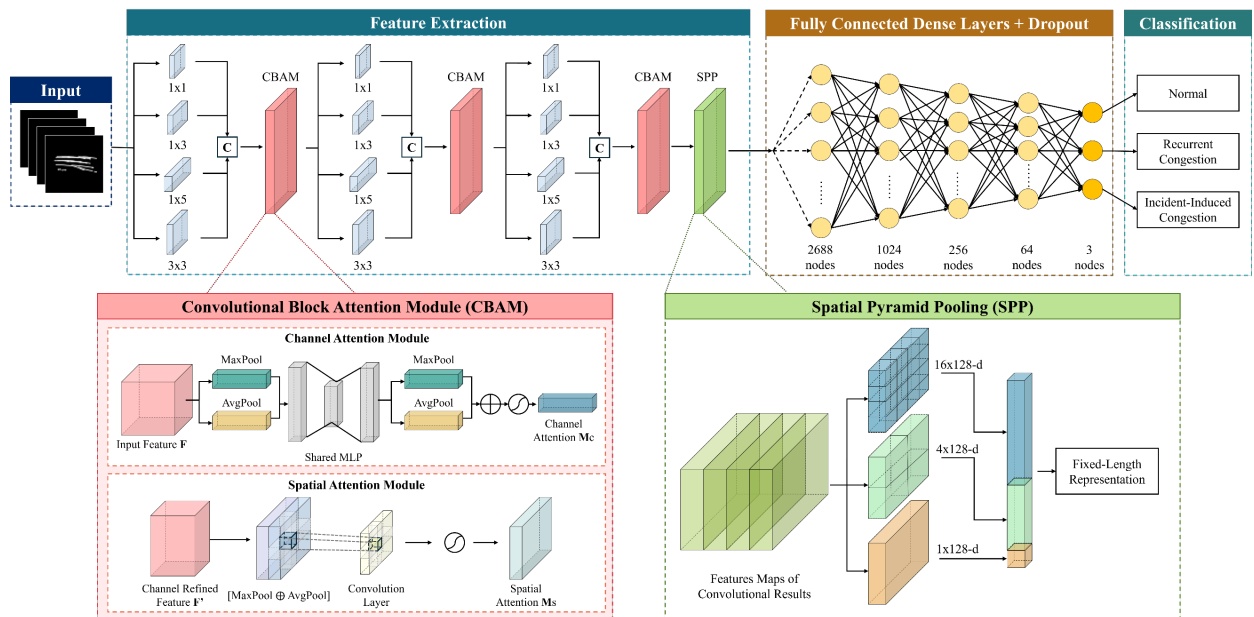

**Figure 3 The architecture of Traffic Condition Detection Network (TCD-Net).**

*Image-to-Video Aggregation*

After classifying all extracted vehicle trajectory images within a video segment, accurately determining whether the segment has captured an incident, recurrent congestion or normal traffic is crucial. To address this, we developed an image-to-video detection result aggregation method that converts image-level detections into a video-level incident detection result.

Specifically, we implemented a statistical aggregation approach by computing the proportion of images within each video segment that are classified as indicating an incident, recurrent congestion or normal traffic (**Figure 1**). For each video segment, we calculate the proportions of extracted vehicle trajectory images classified as incident, recurrent congestion, and normal traffic, denoted by $\tau_I$, $\tau_R$, and $\tau_N$, respectively. Subsequently, if $\tau_I$ exceeds a predefined threshold, the video segment is deemed to have captured a traffic incident or its induced congestion; otherwise, if $\tau_N$ exceeds a predefined threshold, the segment is considered to represent normal traffic conditions. If neither condition is met, the video segment is determined to have captured recurrent congestion. Furthermore, this method effectively mitigates the influence of isolated misclassifications of trajectory images, thereby enhancing the accuracy and robustness of the video-level traffic incident detection.

*Incident Feature Extraction*

After detecting all video segments, extracting incident features including incident scene, incident-induced congestion length and propagation speed is crucial for TMCs to promptly assess incident type, severity, and its impact on traffic. For computing the congestion length and propagation speed, since the drone transmits GPS coordinates to the workstation every second, the spatial range covered by each video segment can be calculated from its corresponding time span. The overall length of the incident-induced non-recurrent congestion is then calculated based on the total GPS coordinate range spanned by consecutive incident-detected video segments, while the congestion propagation speed is obtained by differentiating the change in congestion length over the time interval between two consecutive incident detection flights. Regarding incident scene extraction, since the drone's flight direction aligns with the traffic flow and congestion occurs only upstream of the incident while traffic downstream remains smooth, the incident scene should be captured within the last video segment of a series of consecutively incident-detected video segments in a single detection flight. Within this segment, the precise incident scene time period corresponds to the time interval covered by the longest consecutive trajectory images classified as incidents, since vehicles close to the incident scene exhibit the most frequent lane-changing behavior, the slowest speeds, and the highest density.

**Traffic Incident Detection Platform**

After real-time traffic incident detection and feature extraction are completed, timely visualization of this information to TMCs is vital for facilitating rapid online verification, severity assessment, and effective incident management. Accordingly, this study developed a traffic incident detection platform that deploys the detection framework and visualizes incident detection and feature extraction results in real time through a web-based GUI. The following sections detail the platform's hardware architecture and software implementation.

*Hardware Architecture*

**Figure 4** illustrates the hardware architecture of the platform, which comprises five main components: (1) the drone, (2) the ground control station, (3) the live deck (or an alternative second remote controller), (4) the workstation, and (5) the video live-streaming platform. Regarding signal connectivity, bidirectional communication is established between the drone and both the ground control station and the live deck via a 2.4 GHz wireless link. Communication between the live deck and both the workstation and the video live-streaming platform is facilitated through unidirectional cellular network signals, while the video live-streaming platform and the workstation also communicate unidirectionally via network connections (Wi-Fi, cable, or cellular).

In terms of data transmission, bidirectional data transfer occurs between the drone and the ground control station. The drone transmits real-time thermal camera video and timestamped GPS coordinates to the ground control station, which in turn sends real-time control commands and flight mission plans to the drone. To decouple the operations related to the drone's flight control from thermal video and GPS data transmission, the live deck is employed as a mediator. The live deck, an information-receiving device provided by the drone manufacturer (or alternatively, a second remote controller), communicates with the drone over long distances via a 2.4 GHz wireless link. Information exchange between the live deck and the video live-streaming platform is unidirectional, primarily transmitting thermal camera video from the drone, while the live deck also sends GPS data unidirectionally to the workstation. Concurrently, the workstation accesses the video live-streaming platform to record and store the thermal video data.

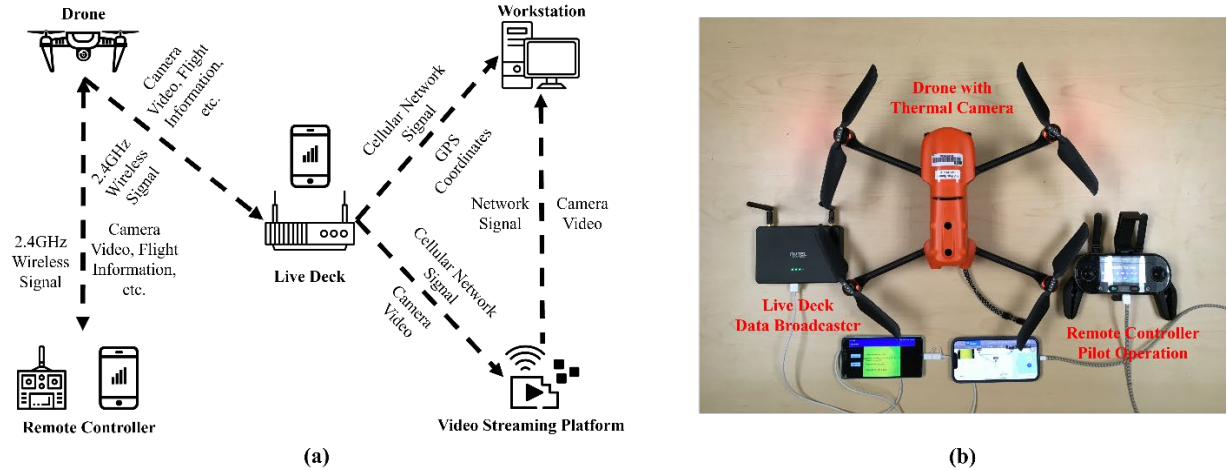

**Figure 4 Hardware of the traffic incident detection platform. (a) Hardware architecture diagram. (b) Physical hardware setup.**

*Software Development*

Building upon the hardware setup, software design is essential to implement the platform's intended functions. This section outlines the software architecture of the platform. As illustrated in **Figure 5**, the software operates concurrently on both the drone side and the workstation side to facilitate data transmission and processing. On the drone side, the software runs on an Android smart device connected to the live deck, enabling the reception and forwarding of thermal video and GPS data transmitted by the drone to the workstation. On the workstation side, the software receives these data and inputs them into the built-in incident detection framework for real-time incident detection and visualization via a web-based GUI.

Specifically, the drone-side software consists of an interface and three threads. The human-computer interaction thread receives operator commands from the interface, which controls the initiation and termination of the other two threads. The thermal video streaming thread transmits the thermal traffic monitoring video from the drone to a video streaming platform, allowing the workstation to capture the live feed. The GPS coordination transmission thread directly transmits timestamped GPS coordinates from the drone to the workstation, ensuring real-time location tracking.

The workstation-side software consists of an interface and five threads. The human-computer interaction thread receives operator commands to manage the initiation and termination of the remaining threads. The GPS coordination receiving thread is responsible for receiving and storing timestamped GPS coordinates from the drone. The thermal video recording thread records the live-streamed video and segments it into 2-minute clips. The traffic incident detection thread uses these 2-minute video segments as input to the incident detection framework, performing real-time incident detection, incident scene time period extraction, and calculation of non-recurrent congestion length and propagation speed. Finally, the detection results visualization thread continuously monitors and visualizes the latest incident detection outcomes on the GUI in real time.

Notably, to enhance the spatial and temporal resolution of incident detection results, the system initiates three thermal video recording threads at 40-second intervals, with each thread recording a 2-minute video segment. Concurrently, three traffic incident detection threads process the corresponding video segments independently. This configuration enables the system to update incident detection results every 40 seconds, significantly improving both spatial and temporal resolution.

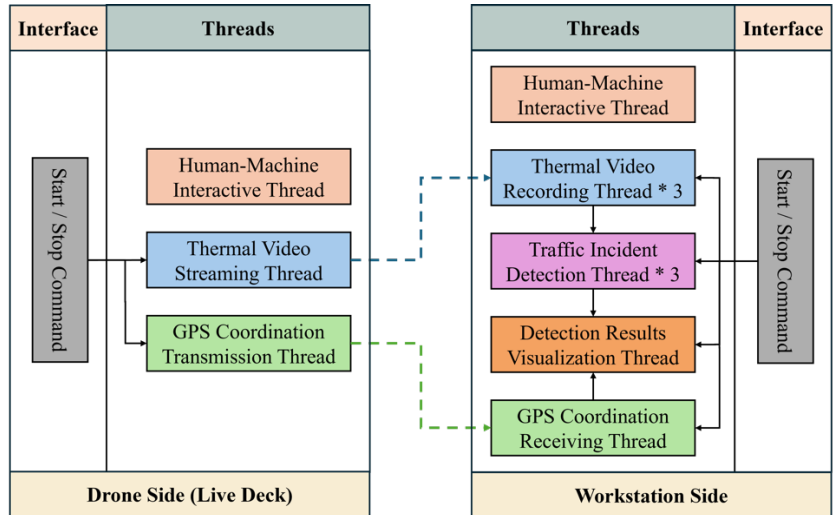
**Figure 5 Software architecture of the traffic incident detection platform.**

**Computing Platform**

  A powerful computing platform can significantly accelerate the training and tuning of deep learning models, thereby enhancing the efficiency of model development and optimization. In this study, an NVIDIA GeForce RTX 3080 GPU was paired with an Intel Xeon Gold 5220R CPU and operated on a Windows 11 system. The DL models were developed, debugged, and executed in a Python 3.10 environment.

**Field Test**

  To validate the accuracy of the traffic incident detection framework and test the functionality of the detection platform, a field test was conducted on the Interstate 75 (I-75) freeway in Wesley Chapel, Florida (**Table 1**). During the test, a rear-end collision occurred on the test road segment, providing a valuable opportunity to evaluate DARTS's incident detection performance.

**Table 1 Timetable and Detailed Information of Test Flights in Field Test**

| Test Road | Road Segment Length | Test Date | Occurrence of Traffic Incidents | Number of Flights | Local Time | Flight Altitude | Cruising Speed |
|---|---|---|---|---|---|---|---|
| Interstate 75 Freeway, Wesley Chapel, FL | 1.4 miles | 2024-04-18 | Yes | 3 | 16:48<br>16:59<br>17:11 | 200 feet | 10 miles per hour |

  It is important to note that, under current Federal Aviation Administration (FAA) regulations, all drones must remain within the operator's Visual-Line-Of-Sight (VLOS) and cannot exceed a maximum flight altitude of 400 feet (*62*). Accordingly, the flight distance in all field tests was restricted to the operator's line of sight, which limited the coverage range of test flights. However, the hardware and software systems developed in this study are designed to support Beyond-Visual-Line-Of-Sight (BVLOS) operations under an FAA waiver (*63*), thereby ensuring scalability for future applications.

**RESULTS**

  Based on the methodology described above, the following investigations were conducted: 1) Drone-based thermal traffic monitoring experiments were performed under incident, recurrent congestion and normal traffic conditions to generate datasets for incident detection DL model training and testing; 2) Regarding the traffic incident detection framework, the effects of different vehicle trajectory extraction periods and image modes on TCD-Net performance were systematically tested. Additionally, the impact of various threshold values in the image-to-video aggregation method on video-level incident detection accuracy was assessed; 3) The software system was developed on both the drone and workstation sides, enabling a web-based interactive GUI that provides real-time visualization of incident detection results, incident scene time periods, and incident-induced congestion length and propagation speed. 4) Field test results were analyzed and benchmarked against existing local TMC incident protocols to evaluate the performance of DARTS in real-time traffic incident detection. The following sections present a detailed description of the results and findings.

**Drone-Based Thermal Traffic Monitoring Dataset**

During the drone-based thermal traffic monitoring process, a total of 64 thermal videos were captured from nine locations on Interstate 4 (I-4), Interstate 75 (I-75), and Interstate 275 (I-275) freeways in the Tampa Bay area of Florida. After organizing and cleaning the video data and excluding invalid videos caused by drone flight parameters not meeting predetermined parameters and blurred video quality, a total of 31 videos met the criteria to be used as the dataset for TCD-Net training and testing. The videos, with durations ranging from 30 to 170 seconds and a cumulative duration of 58.33 minutes, encompassed incident, recurrent congestion and normal traffic conditions. Specifically, 12 videos recorded traffic flow conditions upstream and downstream of incidents, 3 videos recorded upstream and downstream traffic flow under recurrent congestion conditions, while the remaining 16 depicted normal traffic conditions.

In generating the vehicle trajectory image dataset, variations in video durations made it challenging to partition the data into training, validation, and test sets based solely on video count, as this could lead to inconsistencies in image quantity and label distribution across the sets. Considering that the incident detection framework relies on detecting traffic dynamics from vehicle trajectory images extracted at fixed periods, without requiring a strict chronological order, the study first extracted and labeled trajectory images from all videos. Each trajectory image was assigned a label of 2 if its extraction period overlapped with an incident period, 1 if its extraction period overlapped with a recurrent congestion period, and 0 under normal traffic conditions. Stratified sampling was then employed to partition the dataset into training, validation, and test sets in a 6:2:2 ratio, ensuring that the proportions of each label remained consistent. This approach ensures that TCD-Net learns representative features, contributing to more robust training outcomes.

**Table 2 Dataset Size and Label Distribution of Vehicle Trajectory Images Across Different Extraction Periods**

| Trajectory Extraction Period | Number of Labels 2, 1, and 0 | Training Set Size | Validation Set Size | Test Set Size |
|---|---|---|---|---|
| **3s** | 1711 / 128 / 1592 | 2058 | 686 | 687 |
| **5s** | 1689 / 127 / 1553 | 2021 | 674 | 674 |
| **10s** | 1634 / 122 / 1458 | 1928 | 643 | 643 |
| **15s** | 1579 / 117 / 1363 | 1835 | 612 | 612 |
| **20s** | 1524 / 111 / 1269 | 1742 | 581 | 581 |

Furthermore, as previously mentioned, the study examined the effects of five different vehicle trajectory extraction time periods (3 s, 5 s, 10 s, 15 s, and 20 s) and two image modes (monochrome and color) on model performance. Consequently, five distinct datasets were generated, each corresponding to a different extraction period and containing both monochrome and color images **(Table 2)**. Based on subsequent model testing, the extraction time period and image mode yielding the highest accuracy on the test set were selected for integration into the incident detection platform.

**Traffic Incident Detection Framework**

After compiling the dataset, we evaluated the performance of TCD-Net across five datasets and two image modes and assessed the impact of different thresholds on the accuracy of the image-to-video aggregation results. The following sections provide detailed results and findings.

*TCD-Net Performance Evaluation*

In the model training process, an early stopping mechanism was implemented to mitigate model overfitting during training. Specifically, training was terminated if the validation loss did not improve for twenty consecutive epochs beyond the minimum loss, and the model parameters from the epoch with the lowest loss were retained. Additionally, after training different TCD-Nets on five datasets of different trajectory extraction periods and two image modes, we compared their performances to select the one that achieved the highest accuracy for integration into the incident detection platform.

To compare TCD-Net's performance on the trajectory image three-class classification task across different datasets, we adopted six widely used evaluation metrics: loss, accuracy, precision, recall, F1-score, and Area under the Receiver Operating Characteristic Curve (AUC-ROC). Loss serves as the training objective, quantifying the discrepancy between predicted outputs and the true labels. Accuracy represents the overall proportion of correctly classified samples, offering a general performance overview. Precision focuses on the proportion of true positives among the samples predicted as positive for each class, whereas recall measures the proportion of actual positives that are correctly identified. The F1-score, computed as the harmonic mean of precision and recall, balances these two metrics. Finally, the AUC-ROC evaluates the model's discriminative ability by examining the relationship between the true positive rate and the false positive rate across various thresholds. Given that these metrics are well-established in the academic community, only a brief description is provided here without detailing their specific formulas.

**Table 3 Performance of Trained TCD-Nets on Test Sets Under Different Combinations of Trajectory Extraction Periods and Image Modes**

| Image Mode | Trajectory Extraction Period | Loss | Accuracy | Precision | Recall | F1-Score | AUC-ROC |
|---|---|---|---|---|---|---|---|
| monochrome | 3s | 0.0025 | 0.9767 | 0.8928 | 0.9343 | 0.9114 | 0.9959 |
| | 5s | 0.0027 | 0.9777 | 0.9348 | 0.8642 | 0.8947 | 0.9966 |
| | 10s | 0.0040 | 0.9823 | 0.9077 | 0.8986 | 0.9031 | 0.9953 |
| | 15s | 0.0021 | 0.9902 | 0.9547 | **0.9547** | 0.9547 | 0.9976 |
| | **20s** | **0.0013** | **0.9914** | **0.9643** | 0.9519 | **0.9580** | **0.9994** |
| color | 3s | 0.0038 | 0.9694 | 0.8677 | 0.8677 | 0.8677 | 0.9929 |
| | 5s | 0.0036 | 0.9688 | 0.9045 | 0.7896 | 0.8256 | 0.9931 |
| | 10s | 0.0050 | 0.9767 | 0.9055 | 0.8609 | 0.8804 | 0.9882 |
| | 15s | 0.0044 | 0.9624 | 0.8357 | 0.7820 | 0.8026 | 0.9936 |
| | 20s | 0.0028 | 0.9862 | 0.9341 | 0.9341 | 0.9341 | 0.9971 |

**Table 3** presents the performance of the trained TCD-Net on the test set under different combinations of trajectory extraction periods and image modes, quantitatively evaluated using six performance metrics. The results indicate that TCD-Net achieves superior performance in all six metrics when using the monochrome image mode compared to the color image mode for datasets with the same extraction period. This phenomenon suggests that, for this task, the model primarily relies on the edge and structural features of vehicle trajectories for classification; monochrome images could more directly highlight the shape and spatial layout of trajectories, while color information may introduce extraneous noise that interferes with key feature extraction. Regarding the impact of the extraction period on model performance, the results show that the TCD-Net trained on the dataset with a 20-second extraction period attains the best classification performance. This finding indicates that a longer extraction period can capture more vehicle trajectories and better reflect the differences in trajectory length and distribution, thereby providing richer spatiotemporal features that enhance the model's ability to differentiate traffic conditions. Finally, we deployed the TCD-Net model trained with the monochrome image mode and a 20-second trajectory extraction period as the DL model for the traffic incident detection platform, and this configuration will serve as the standard for vehicle trajectory image generation in subsequent field tests.

*Thresholds for Image-to-Video Aggregation*

The TCD-Net detects each trajectory image to determine whether it indicates a traffic incident or its induced congestion. To generate a trinary detection result for each video segment, the detection results of all trajectory images within that segment are aggregated. Specifically, for each video segment, we calculate the proportions of extracted vehicle trajectory images classified as incident, recurrent congestion, and normal traffic, denoted by $\tau_I$, $\tau_R$ and $\tau_N$, respectively, and classify the segment based on whether these ratios exceed their corresponding thresholds. This approach consolidates image-level detections while mitigating the impact of isolated misclassifications on the overall video-level result.

To determine the optimal thresholds, we set the incident and normal thresholds to values ranging from 0.1 to 0.9, forming 81 different threshold combinations. These combinations were then applied for image-to-video aggregation on 31 videos in the dataset. Finally, we recorded the video classification accuracy and the corresponding number of correctly classified videos for each combination, as shown in **Table S1**. Experimental results indicate that these 81 threshold combinations yielded video classification accuracies ranging from 97% to 100%, thereby demonstrating the effectiveness of the image-to-video aggregation method. In the specific incident detection task, to improve detection sensitivity, we selected the most stringent combination among those achieving 100% accuracy, namely, an incident threshold of 0.1 and a normal threshold of 0.6, as the standard configuration for the image-to-video aggregation method in the incident detection platform.

**Traffic Incident Detection Platform**

After optimizing the key components of the framework, developing the traffic incident detection platform became a critical step in implementing DARTS. Building upon the established hardware and software designs, we developed dedicated software for both the drone and workstation sides, tailored to the requirements of drone operators and TMC personnel. To balance hardware constraints with user accessibility, the drone-side software was implemented as a native Android application, while the workstation-side software was realized as a web-based application.

*Drone Side Software*

The Android software on the drone side is responsible for forwarding the thermal video and drone GPS coordinate data from the drone to both the video streaming platform and workstation. **Figure S1(a)** is the main interface, which employs a grid layout to display various operational modules, each representing a distinct function. **Figure S1(b)** shows the live streaming interface accessed by clicking the "NICR Live Push" button. It displays a confirmation message for a successful connection along with essential streaming details such as the date, time, frame rate, and audio bit rate. The lower section of the status message indicates the transmission status of the drone's GPS data to the workstation. Additionally, two buttons on the left enable the operator to manually control the start and stop of live streaming and GPS data transmission. **Figure S1(c)** depicts the same interface when both live streaming and GPS data transmission have been halted, as indicated by the displayed status messages.

*Workstation Side Software*

The web-based application on the workstation side is responsible for receiving GPS coordinates and thermal video data transmitted by the drone-side software. It executes the traffic incident detection framework on the backend to detect traffic incidents, extract the incident scene time period and estimate the length and propagation speed of incident-induced congestion. The detection results are then visualized in real time on the webpage, enabling TMC personnel to verify and monitor incidents effectively. Additionally, the application processes operator commands to start and stop the detection process.

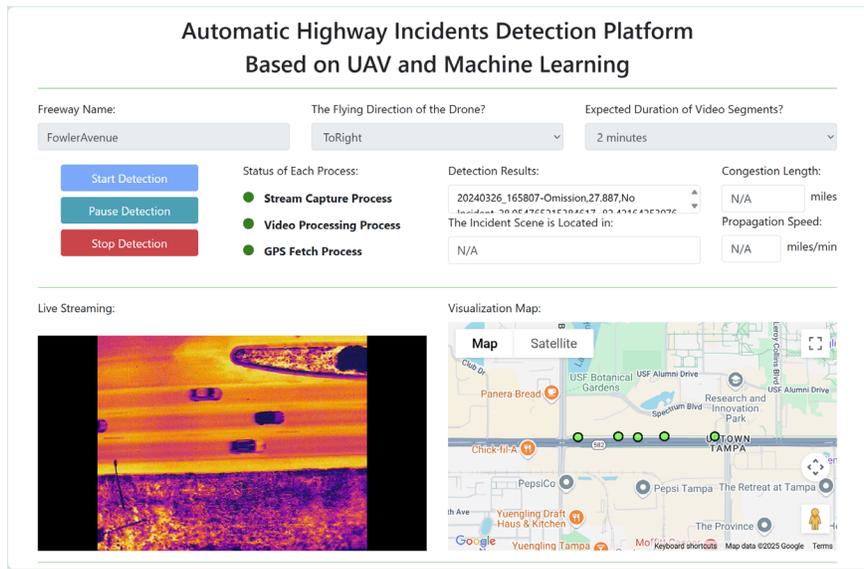

**(a)**

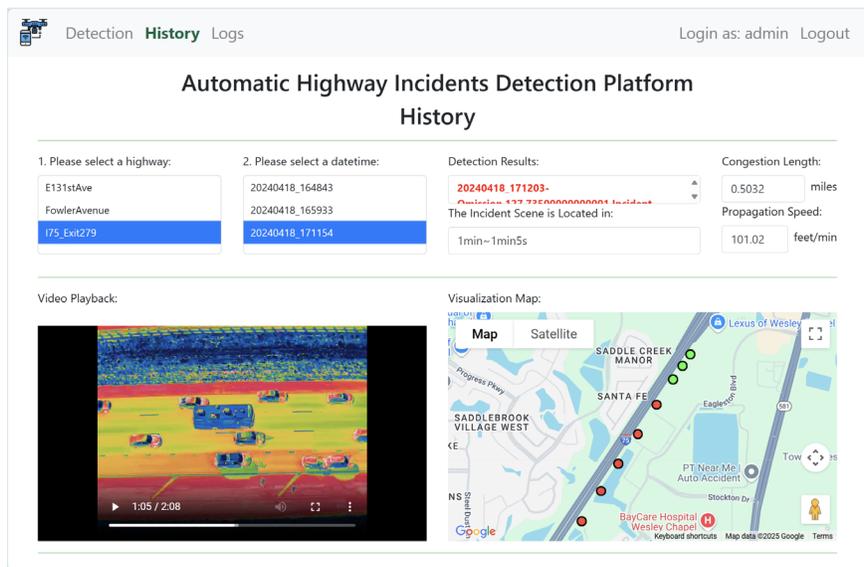

**(b)**

**Figure 6 Webpages for detection and history review in the traffic incident detection platform. (a) The detection webpage during a detection process. (b) The history review webpage.**

    To support detection task control, results visualization, and review of historical detections, two dedicated webpages, which are detection and history review webpages, were developed to ensure efficient operation and accurate presentation (**Figure 6**). The detection webpage facilitates the initiation, control, and real-time visualization of incident detection tasks. It displays a live thermal video stream alongside formatted detection results in a dedicated textbox and presents incident locations on an integrated visualization map. Control buttons allow operators to start, pause, or stop the detection process, while additional fields show metrics such as congestion length and propagation speed, as well as the incident scene's time period for detailed analysis. The history review webpage provides an overview of past incident detection flights by allowing operators to select a freeway and a specific date/time to access

stored detection results and thermal video segments. Detection results are presented in both textual and visual formats, enabling operators to replay the corresponding video segments and review historical traffic conditions efficiently. On the visualization maps of these two pages, incident-induced congestion, recurrent congestion, and normal traffic conditions are represented by red, orange, and green dots, respectively. To ensure optimal visualization across various devices, the web-based GUI employs a responsive design that adapts to both desktop and mobile screen sizes (**Figure S2**).

Finally, dedicated web pages were also developed for GUI access control and system debugging. For operational security, the application features a login page to protect critical system functions. **Figures S3(a)** and **S3(b)** display screenshots of the login page on different devices. Operators must authenticate to access system functions. Moreover, a log webpage displays real-time backend thread logs to help operators monitor system status (**Figure S4**).

**Field Test**

After developing and optimizing the traffic incident detection framework and platform, multiple field tests were conducted to evaluate the detection accuracy and performance of DARTS. During the field test on April 18$^{th}$, a rear-end collision involving two vehicles occurred, providing a valuable opportunity to assess DARTS for real-time traffic incident detection, incident scene time period extraction, and calculation of incident-induced non-recurrent congestion length and propagation speed. Three consecutive test flights on April 18$^{th}$ successfully captured the changes in road traffic flow conditions before and after the incident, incident scene time period, and the propagation of the traffic-induced non-recurrent congestion (**Figure 7**). The following sections detail DARTS's performance in traffic incident detection during the field test and compare it with the activities of local TMCs recorded on traffic information platforms during the incident period.

*Incident Detection Results from DARTS*

The historical review web pages of the three detection flights in the DARTS field test are shown in **Figure 7**. It can be observed from the figure that during the first flight, the system detected recurrent congestion at the off-ramp of I-75 Exit 279 (**Figure 7(a)**). Since this flight took place during peak traffic hours and the congestion was concentrated at the off-ramp, downstream of which an intersection controlled by a traffic signal exists, it was inferred that the congestion was caused by a traffic signal restricting traffic flow. This finding suggests signal timing inadequacies that failed to adapt to fluctuating traffic demands and indicates that such bottleneck congestion may serve as a precursor to subsequent traffic crashes.

In the second flight, the system detected an incident-induced non-recurrent congestion pattern, with three detection points upstream of the off-ramp labeled as indicating incident-induced congestion (**Figure 7(b)**). Moreover, the webpage displayed that the current incident-induced non-recurrent congestion extended for 0.265 miles. Additionally, the visualized detection results on the map showed that the recurrent congestion previously detected on the off-ramp had dissipated. In the third flight, the system continued to detect an incident-induced non-recurrent congestion pattern, and at that time, all detection points upstream of the off-ramp were marked as incident-induced congestion (**Figure 7(c)**). This indicates that the system not only detected the traffic incident but also captured the upstream propagation process of the incident-induced congestion. The webpage also reported that in the third flight, the detected incident-induced congestion spanned 0.5032 miles, propagating upstream at a speed of 101.02 feet per minute. Finally, the webpages for both the second and third flights provided the time period during which the incident scene—extracted by the incident detection framework—appeared in the first upstream incident-detected video segment. Based on this cue and the video playback functionality offered on the webpage, we successfully pinpointed the timestamp of the incident scene and confirmed the presence, type, and severity of the traffic incident in real time. Furthermore, according to the start time of the second flight, the drone's detection flight speed, and the timestamped traffic incident detection records shown on the webpage, we determined that the system identified the crash at approximately 5:03 PM.

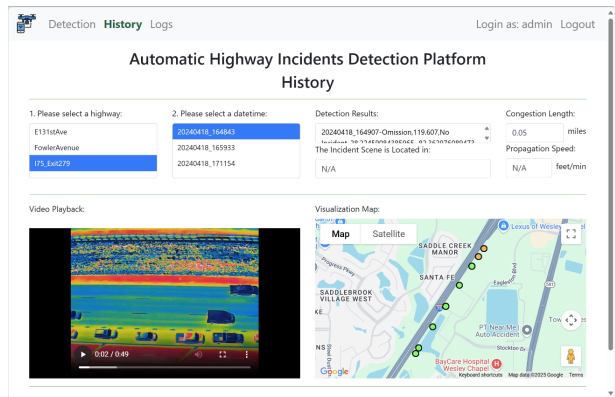
(a)

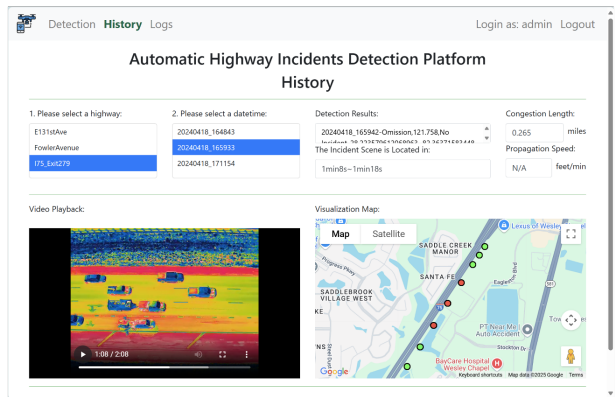
(b)

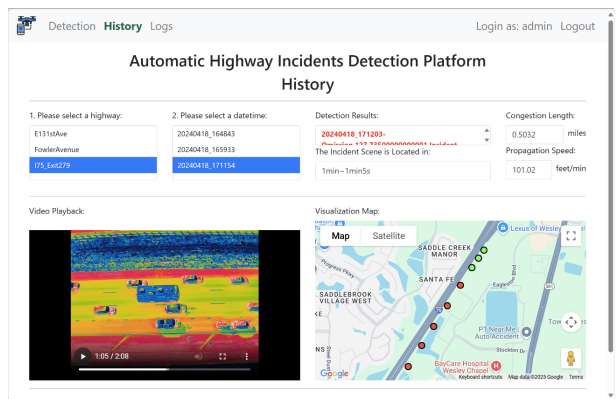
(c)

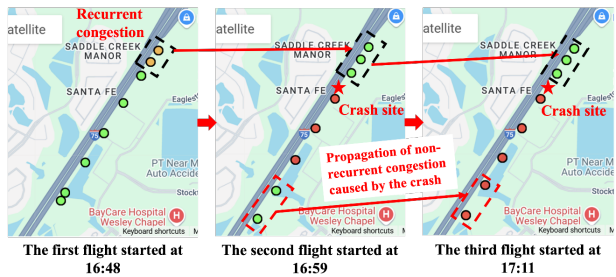
(d)

**Figure 7** Traffic incident detection results from three detection flights. (a) The first flight started at 16:48. (b) The second flight started at 16:59. (c) The third flight started at 17:11. (d) The propagation process of incident-induced non-recurrent congestion captured during the field test.

*Comparison with Traffic Information Platforms*

To compare with the activities of local TMCs recorded on traffic information platforms during the same incident period, we obtained incident records from the Florida 511 website (FL511) and the Regional Integrated Transportation Information System (RITIS) for this crash. The FL511 data were used to verify the crash location and approximate start time (**Figure 8**). **Figure 8(a)** shows the crash site and its induced congestion information detected by DARTS during the field test, while **Figure 8(b)** presents the location of the rear-end collision as reported by TMC and road rangers on the FL511 platform. **Figure 8(c)** displays a surveillance image of the collision, which involved a red sedan rear-ending an SUV. According to FL511, this two-vehicle collision occurred just before Exit 279 on I-75 at approximately 5:09 PM.

The RITIS records provide more detailed, accurate, and comprehensive information. **Figure 9** illustrates the entire process of the crash, from the initial report to verification and eventual resolution, along with the corresponding key timestamps. The RITIS screenshot indicates that the crash was reported to the Florida Highway Patrol at approximately 5:05 PM and was subsequently verified by the Florida Department of Transportation (FDOT) around 5:15 PM. Additionally, the crash resulted in non-recurrent congestion extending more than 5 miles upstream, persisting until approximately 7:00 PM.

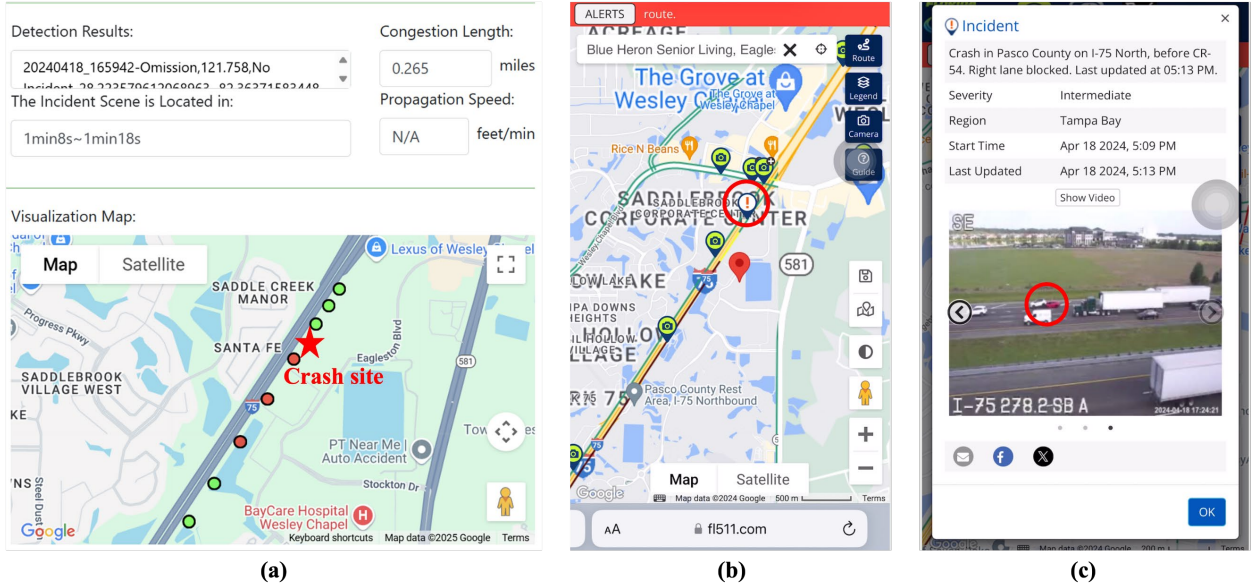

Figure 8 Traffic incident information. (a) The detection result from the DARTS. (b) The screenshot of the incident on FL511. (c) The incident information on FL511.

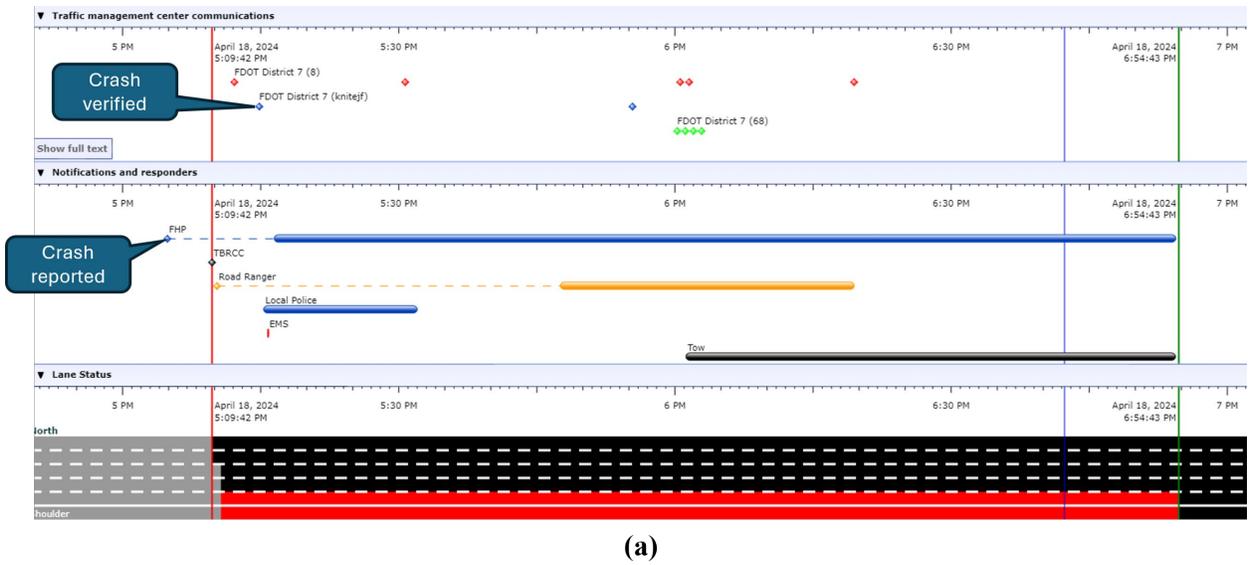

Figure 9 Activities of local TMCs recorded on RITIS. (a) System interface screenshot. (b) Detailed records of local TMC activities.

In contrast, our system detected the crash at 5:03 PM, enabling TMC personnel to directly verify the incident and assess its severity by reviewing the extracted incident scene and video playback on the webpage. This demonstrates that the system detected the crash 12 minutes earlier than the local TMC. By providing timely detection and immediate online verification, the system facilitates prompt medical assistance for victims and swift incident management, thereby enhancing response efficiency and mitigating delays associated with traditional incident verification, where road rangers may be delayed or unable to access the incident site due to congestion. Moreover, although the timeliness of drone-based incident detection is closely related to its distance from the incident site, the drone system designed in this study is intended to conduct continuous patrols along freeways to detect traffic incidents actively. Furthermore, the deployment of a drone fleet patrolling different freeway segments could further reduce detection delays.

**DISCUSSION**

This study presents significant advancements in traffic incident detection. The following sections elaborate on additional contributions and perspectives, including the creation of original datasets, the scalability and adaptability potential of DARTS, and its anticipated contributions to global road safety and sustainable transport objectives.

**First Drone-Based Thermal Traffic Monitoring Dataset**

Addressing the nascent stage of drone-based traffic incident detection research (*52*), a critical barrier has been the lack of publicly available, specialized datasets. Between 2021 and 2023, this study undertook extensive data collection campaigns across nine distinct locations along freeways (I-4, I-75, I-275) in Florida's Tampa Bay area. Utilizing drone-mounted thermal cameras, these efforts yielded a unique collection of thermal video data capturing diverse traffic conditions, including incidents, recurrent congestions, and normal traffic. A total of five vehicle trajectory image datasets were generated, with each dataset corresponding to a distinct extraction period and comprising between 2,904 and 3,431 labeled images. As no comparable public datasets currently exist, this collection represents the first publicly available resource specifically for drone and thermal camera-based traffic incident detection. It provides an essential foundation, enabling further research and development in this emerging and critical field by offering testbed data for algorithm validation and comparative analysis.

**Perspective of a Distributed Multi-Drone Patrolling System**

A common limitation of existing incident detection systems lies in their deployment rigidity and operational inefficiencies. The framework developed in this study demonstrates notable computational efficiency, requiring only modest onboard resources. This feature suggests the potential for running the detection framework directly on edge computing devices mounted on drones (*64*), enabling onboard data processing. Such a capability lays the technical groundwork for future implementations of distributed, multi-drone systems dedicated to autonomous patrolling and incident detection.

The practical feasibility of this vision is further supported by evolving regulatory frameworks, such as the FAA's established procedures for BVLOS operations (*63*), which are critical for enabling fully autonomous drone patrols on public roads. In parallel, advances in supporting infrastructure, most notably, the development of drone docking stations, present a promising pathway for scalable deployment. Commercially available docking systems from manufacturers like DJI and Skydio already offer automated takeoff, landing, and battery charging capabilities in remote settings (*65, 66*). These platforms are also designed for resilience in extreme weather (e.g., hurricanes, blizzards) and may facilitate rapid redeployment for post-event incident detection when ground access is compromised.

Compared to traditional fixed surveillance systems such as CCTV, drone-based platforms offer enhanced operational flexibility. Patrolling schedules and routes could be dynamically adapted based on real-time traffic conditions or shifting incident hotspots, providing more targeted monitoring where and when it is most needed. While the current implementation focuses on a single-drone prototype, the modular design of DARTS lends itself to future multi-drone coordination, which may enable broader area coverage or intensified surveillance of high-risk zones—capabilities that stationary systems inherently lack.

The ability of a single drone to patrol a relatively large area per flight cycle also suggests the feasibility of scalable deployment through coordinated multi-drone operations. Preliminary estimates suggest that, operating at a cruising speed of 10 mph, a single drone with a flight duration of approximately 38 minutes could cover a patrol distance of around 5 miles per battery cycle. This potential range is notably greater than the typical 0.5-mile effective coverage of fixed CCTV installations (*67*, *68*), indicating a promising opportunity to improve surveillance efficiency and asset utilization. While these figures are based on initial assumptions, they inform future designs of distributed multi-drone systems covering wider freeway networks or remote regions.

From an operational perspective, executing the detection framework directly onboard drones reduces the computational load at TMCs, enabling a decentralized architecture that requires only modest server capacity for coordinating fleets, receiving incident alerts, and visualizing data via the DARTS interface. This contrasts with centralized systems that depend on high-performance computing infrastructure. The modular, lightweight design of DARTS supports cost-efficient deployment and offers particular promise in regions with limited infrastructure or coverage gaps, including remote and underserved areas. As such, the system has the potential to enhance road surveillance capabilities across diverse urban and rural settings, while lowering reliance on frequent physical patrols. These features suggest that DARTS could contribute to more inclusive and scalable incident management strategies, making advanced detection technology more accessible in resource-constrained environments.

**CONCLUSIONS**

Addressing the critical limitations inherent in conventional traffic incident detection methods, namely delays in detection, lack of online visual verification and severity assessment, difficulties in assessing non-recurrent congestion impacts, and operational constraints under low-visibility conditions or privacy concerns, this study introduced an innovative drone-based, AI-powered system, DARTS. By synergistically integrating the high mobility of drones, the low-visibility operation and privacy-preserving capabilities of thermal cameras, and a customized, lightweight deep learning model (TCD-Net), this research developed a framework capable of continuously extracting vehicle trajectories and detecting incidents from aerial thermal video streams. This framework enables real-time detection of traffic incidents and associated non-recurrent congestion, while critically distinguishing these from routine, recurrent congestion patterns.

The culmination of this research is DARTS, a fully integrated hardware-software platform featuring an interactive GUI, designed and implemented for practical deployment by TMCs. This system directly bridges the identified research gap by providing simultaneous capabilities for 1) rapid, real-time incident detection; 2) immediate online visual verification and preliminary severity assessment via thermal video feed; and 3) dynamic monitoring of the impact range and propagation of incident-induced non-recurrent congestion. The development represents a complete closed-loop system, streamlining the workflow from automated incident detection through verification to actionable reporting for TMCs.

Field testing conducted on the I-75 freeway in Florida provided compelling validation of the system's efficacy. DARTS successfully detected a rear-end collision and accurately tracked the resulting non-recurrent congestion and its upstream propagation dynamics. Crucially, benchmarking against official RITIS records revealed that DARTS achieved incident detection 12 minutes earlier than the local TMC, simultaneously providing immediate online access to the incident scene for verification. These findings underscore the system's potential to accelerate emergency response times, thereby mitigating the risk of increased injury severity or fatalities often associated with delays in emergency medical services. Furthermore, the demonstrated ability to monitor congestion propagation offers TMCs valuable data for proactive traffic management interventions, aimed at reducing congestion propagation and preventing secondary incidents. The system also successfully identified recurrent congestion patterns prior to the incident, confirming its capability to differentiate between incident-induced and regular traffic flow disruptions.

In essence, this study lays a solid foundation for future robotic patrolling applications in roadway traffic management. The drone-based approach demonstrated here offers notable flexibility, enabling adaptive deployment that could complement or enhance fixed-location surveillance infrastructure, particularly in scenarios where incident hotspots shift dynamically. By integrating real-time detection with immediate online verification and severity assessment, DARTS reduces the need for frequent, resource-intensive on-road patrols, presenting a potentially cost-effective addition to the current incident management method. These capabilities may hold particular relevance for regions with limited access to conventional infrastructure.

As a pilot implementation, the successful deployment and field validation of DARTS demonstrate the feasibility of this approach in a real-world context. While broader deployment remains the goal for future work, the system's performance suggests meaningful potential for scaling to distributed multi-drone operations and integration into next-generation intelligent transportation frameworks. By offering a flexible, infrastructure-light solution that combines real-time detection, online verification, and congestion monitoring, DARTS contributes to the development of more adaptive and efficient traffic incident management strategies. These capabilities are particularly valuable in environments with limited access to conventional surveillance infrastructure, supporting efforts to enhance transportation system resilience, equity, and responsiveness.

**ACKNOWLEDGMENTS**
The authors gratefully acknowledge support provided by the National Institute for Congestion Reduction (NICR), a National University Transportation Centers sponsored by the US Department of Transportation through Grant No. 69A3551947136. The contents of this manuscript reflect the views of the authors, who are responsible for the facts and the accuracy of the information presented herein. The contents do not necessarily reflect the official views or policies of US Department of Transportation.

We would also like to express our gratitude to the Florida Department of Transportation District 7 Traffic Operations office, the Regional Transportation Management Center (RTMC), and the Road Ranger program for permitting the drone pilot to accompany the Road Rangers, enabling data collection in the early stages of this project. Additionally, we acknowledge the Florida 511 website provided by FDOT's RTMCs for offering real-time traffic information that was invaluable throughout this project.

**AUTHOR CONTRIBUTIONS**
**Bai Li**: Methodology, Software, Formal analysis, Investigation, Data Curation, Writing - Original Draft, Visualization.
**Achilleas Kourtellis**: Data Curation, Validation, Investigation, Resources, Writing - Review & Editing.
**Rong Cao**: Software, Formal analysis, Writing - Original Draft, Visualization.
**Joseph Post**: Methodology, Validation, Writing - Review & Editing.
**Brian Porter**: Investigation, Data Curation, Writing - Review & Editing.
**Yu Zhang**: Conceptualization, Methodology, Validation, Resources, Writing - Review & Editing, Supervision, Project administration, Funding Acquisition.

**COMPETING INTERESTS**
The authors declare no competing interests.

**DATA AVAILABILITY**
The data that support the findings of this study and the drone-based traffic incident monitoring dataset are available upon reasonable request from the corresponding author [YZ]. The webpage of DARTS can be accessed via https://detection.sum-lab.duckdns.org with account name of "admin" and password of "usfsumlab".

**DECLARATION OF GENERATIVE AI AND AI-ASSISTED TECHNOLOGIES IN THE WRITING PROCESS**

During the preparation of this manuscript, the author(s) used ChatGPT to check grammar, spelling, and syntax errors.

**Supplementary Information for "*DARTS: A Drone-Based AI-Powered Real-Time Traffic Incident Detection System*"**

**Bai Li[1], Achilleas Kourtellis[2], Rong Cao[3], Joseph Post[1], Brian Porter[4], Yu Zhang[1]\*,**

[1] Department of Civil and Environmental Engineering, College of Engineering, University of South Florida, Tampa, FL, 33620, USA

[2] Center for Urban Transportation Research, College of Engineering, University of South Florida, Tampa, FL, 33620, USA

[3] School of Civil and Environmental Engineering, Nanyang Technological University, 50 Nanyang Ave, 639798, Singapore

[4] District Seven, Florida Department of Transportation, Tampa, FL, 33612, USA

\*Corresponding author: Address as above. Email address: yuzhang@usf.edu

E-mail addresses: baili@usf.edu (B. Li), kourtellis@usf.edu (A. Kourtellis), rong021@e.ntu.edu.sg (R. Cao), japost@usf.edu (J. Post), Brian.Porter@dot.state.fl.us (B. Porter), yuzhang@usf.edu (Y. Zhang).

# SUPPLEMENTARY FIGURES

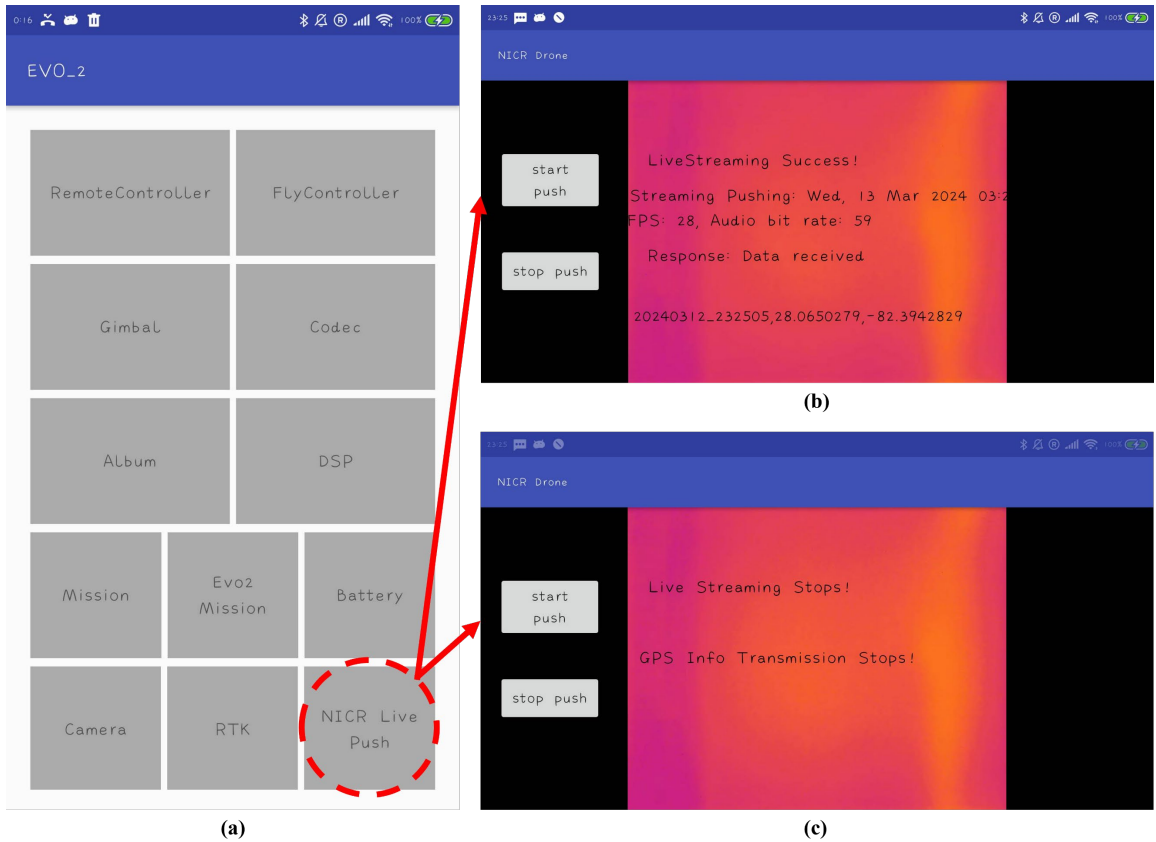

**Figure S1** Android application interfaces on the drone side of the incident detection platform. (a) Main interface. (b) Information forwarding interface during forwarding. (c) Information forwarding interface when forwarding stops.

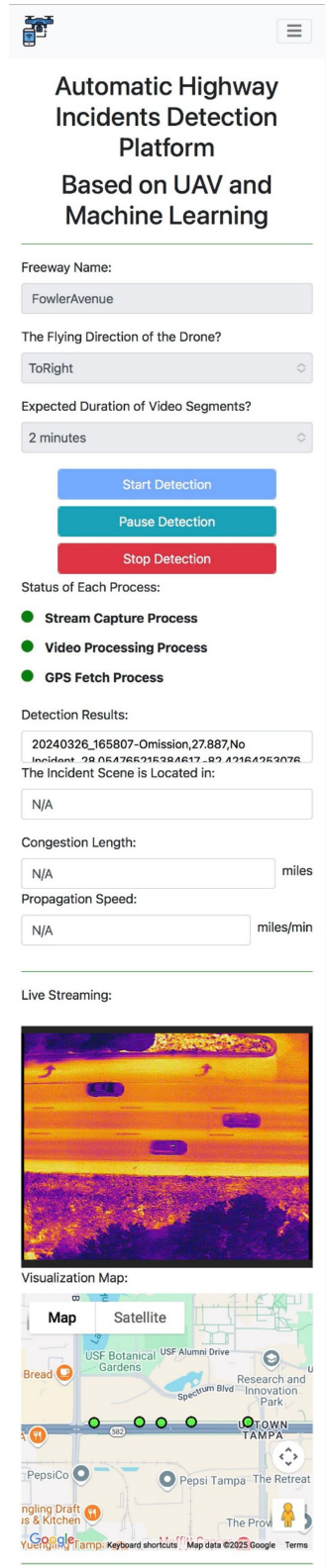 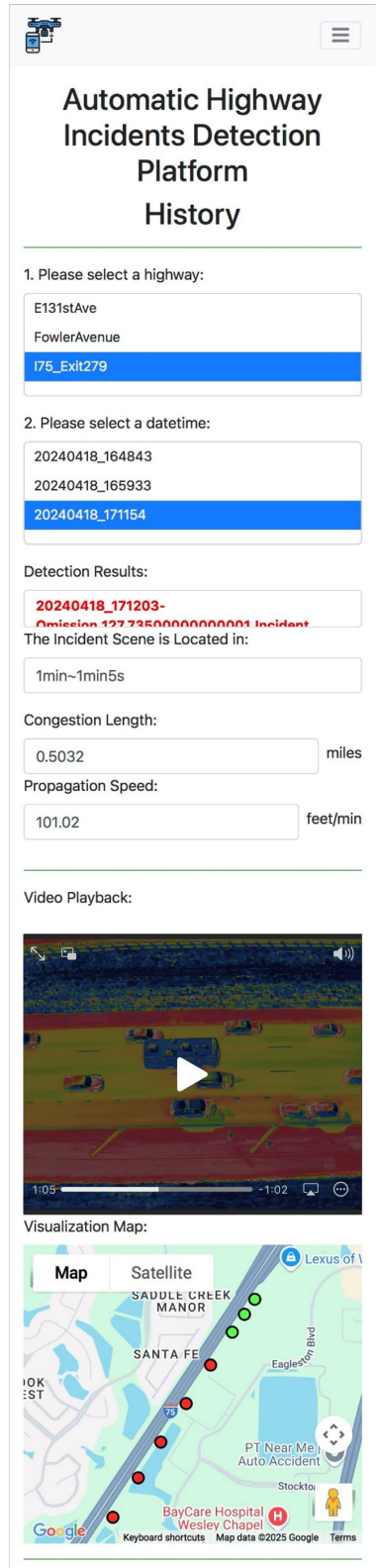

(a)            (b)

**Figure S2 Screenshots of detection and history review webpages on mobile devices. (a) Detection webpage during an ongoing detection process. (b) History review webpage.**

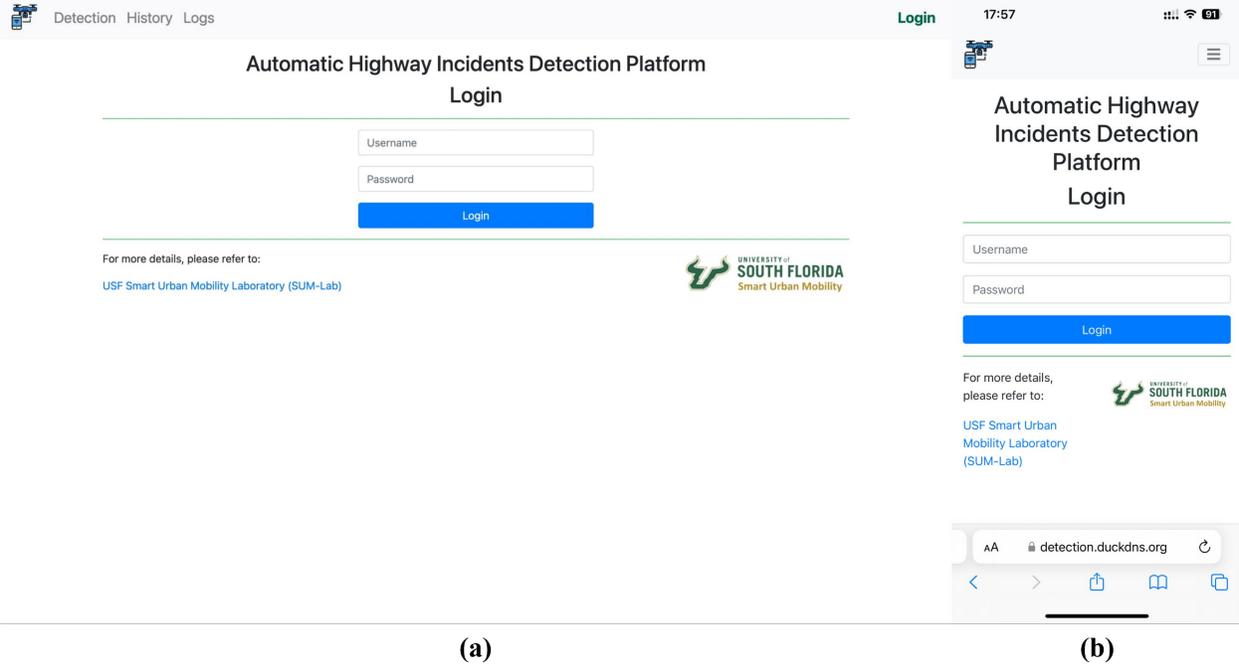

**Figure S3** Login webpages of the traffic incident detection platform. (a) The login webpage on desktop devices. (b) The login webpage on mobile devices

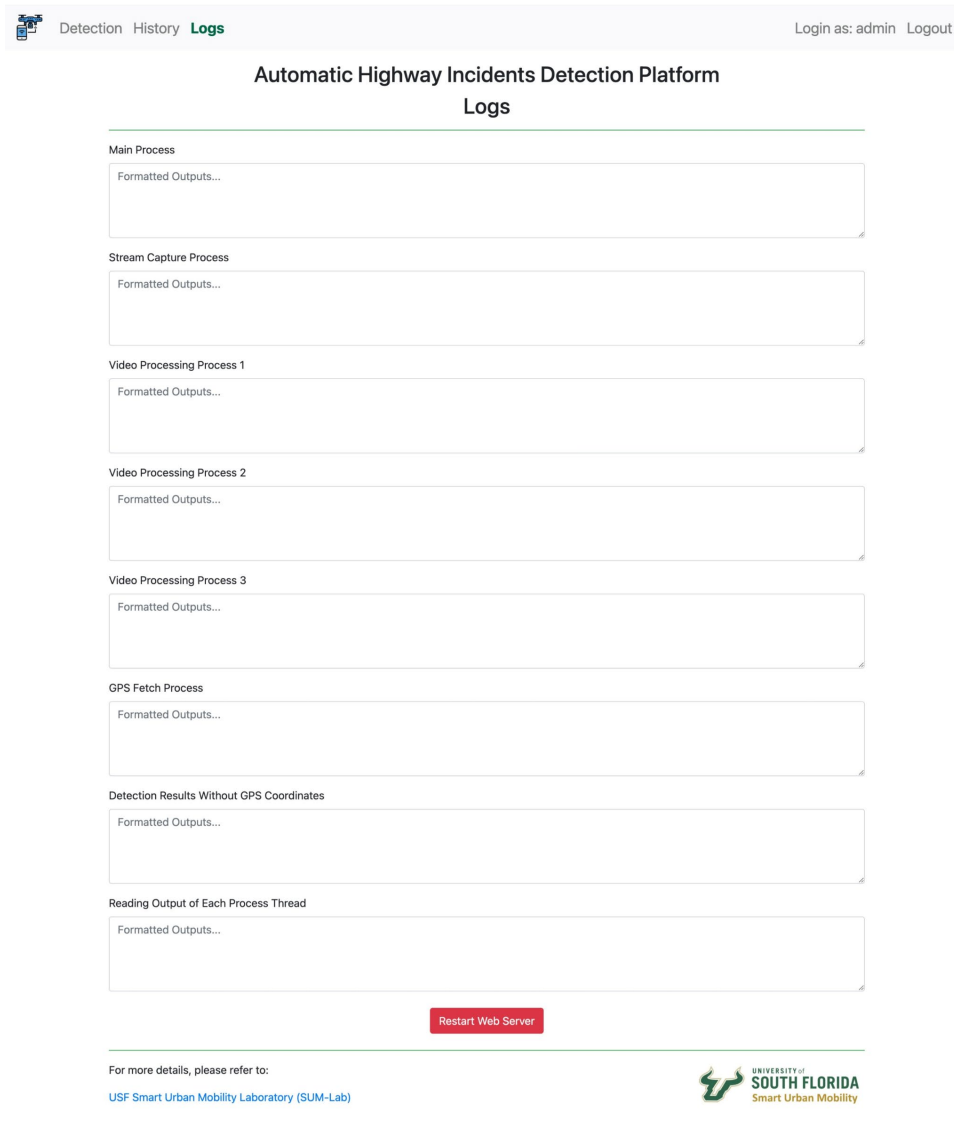
(a)

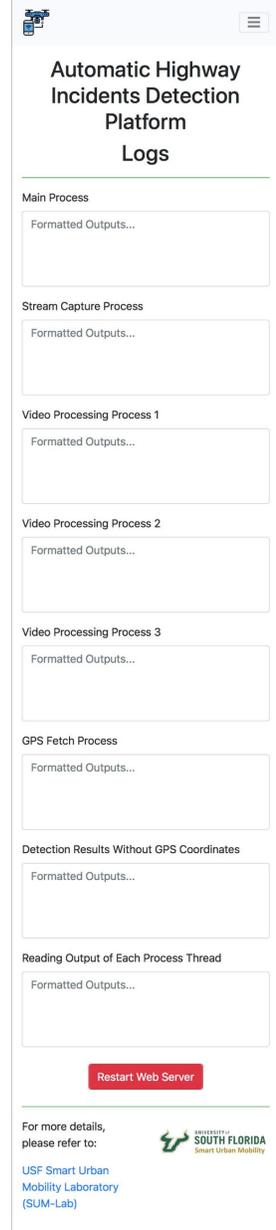
(b)

**Figure S4** Logs webpages of the traffic incident detection platform. (a) The logs webpage on desktop devices. (b) The logs webpage on mobile devices

# SUPPLEMENTARY TABLE

**Table S1 Accuracy Distribution of Image-To-Video Aggregation Across Varying Incident and Normal Classification Thresholds**

| Index | Incident Threshold | Normal Threshold | Accuracy | Correctly Classified Videos |
|---|---|---|---|---|
| 1 | 0.1 | 0.1 | 0.97 | 30 |
| 2 | 0.1 | 0.2 | 0.97 | 30 |
| 3 | 0.1 | 0.3 | 0.97 | 30 |
| 4 | 0.1 | 0.4 | 0.97 | 30 |
| 5 | 0.1 | 0.5 | 0.97 | 30 |
| 6 | 0.1 | 0.6 | 1 | 31 |
| 7 | 0.1 | 0.7 | 1 | 31 |
| 8 | 0.1 | 0.8 | 1 | 31 |
| 9 | 0.1 | 0.9 | 1 | 31 |
| 10 | 0.2 | 0.1 | 0.97 | 30 |
| 11 | 0.2 | 0.2 | 0.97 | 30 |
| 12 | 0.2 | 0.3 | 0.97 | 30 |
| 13 | 0.2 | 0.4 | 0.97 | 30 |
| 14 | 0.2 | 0.5 | 0.97 | 30 |
| 15 | 0.2 | 0.6 | 1 | 31 |
| 16 | 0.2 | 0.7 | 1 | 31 |
| 17 | 0.2 | 0.8 | 1 | 31 |
| 18 | 0.2 | 0.9 | 1 | 31 |
| 19 | 0.3 | 0.1 | 0.97 | 30 |
| 20 | 0.3 | 0.2 | 0.97 | 30 |
| 21 | 0.3 | 0.3 | 0.97 | 30 |
| 22 | 0.3 | 0.4 | 0.97 | 30 |
| 23 | 0.3 | 0.5 | 0.97 | 30 |
| 24 | 0.3 | 0.6 | 1 | 31 |
| 25 | 0.3 | 0.7 | 1 | 31 |
| 26 | 0.3 | 0.8 | 1 | 31 |
| 27 | 0.3 | 0.9 | 1 | 31 |
| 28 | 0.4 | 0.1 | 0.97 | 30 |
| 29 | 0.4 | 0.2 | 0.97 | 30 |
| 30 | 0.4 | 0.3 | 0.97 | 30 |
| 31 | 0.4 | 0.4 | 0.97 | 30 |
| 32 | 0.4 | 0.5 | 0.97 | 30 |
| 33 | 0.4 | 0.6 | 1 | 31 |
| 34 | 0.4 | 0.7 | 1 | 31 |
| 35 | 0.4 | 0.8 | 1 | 31 |
| 36 | 0.4 | 0.9 | 1 | 31 |
| 37 | 0.5 | 0.1 | 0.97 | 30 |
| 38 | 0.5 | 0.2 | 0.97 | 30 |
| 39 | 0.5 | 0.3 | 0.97 | 30 |
| 40 | 0.5 | 0.4 | 0.97 | 30 |
| 41 | 0.5 | 0.5 | 0.97 | 30 |
| 42 | 0.5 | 0.6 | 1 | 31 |
| 43 | 0.5 | 0.7 | 1 | 31 |
| 44 | 0.5 | 0.8 | 1 | 31 |
| 45 | 0.5 | 0.9 | 1 | 31 |
| 46 | 0.6 | 0.1 | 0.97 | 30 |
| 47 | 0.6 | 0.2 | 0.97 | 30 |
| 48 | 0.6 | 0.3 | 0.97 | 30 |
| 49 | 0.6 | 0.4 | 0.97 | 30 |
| 50 | 0.6 | 0.5 | 0.97 | 30 |
| 51 | 0.6 | 0.6 | 1 | 31 |
| 52 | 0.6 | 0.7 | 1 | 31 |
| 53 | 0.6 | 0.8 | 1 | 31 |
| 54 | 0.6 | 0.9 | 1 | 31 |
| 55 | 0.7 | 0.1 | 0.97 | 30 |
| 56 | 0.7 | 0.2 | 0.97 | 30 |
| 57 | 0.7 | 0.3 | 0.97 | 30 |
| 58 | 0.7 | 0.4 | 0.97 | 30 |
| 59 | 0.7 | 0.5 | 0.97 | 30 |
| 60 | 0.7 | 0.6 | 1 | 31 |
| 61 | 0.7 | 0.7 | 1 | 31 |
| 62 | 0.7 | 0.8 | 1 | 31 |
| 63 | 0.7 | 0.9 | 1 | 31 |
| 64 | 0.8 | 0.1 | 0.97 | 30 |
| 65 | 0.8 | 0.2 | 0.97 | 30 |
| 66 | 0.8 | 0.3 | 0.97 | 30 |
| 67 | 0.8 | 0.4 | 0.97 | 30 |
| 68 | 0.8 | 0.5 | 0.97 | 30 |
| 69 | 0.8 | 0.6 | 1 | 31 |
| 70 | 0.8 | 0.7 | 1 | 31 |
| 71 | 0.8 | 0.8 | 1 | 31 |
| 72 | 0.8 | 0.9 | 1 | 31 |
| 73 | 0.9 | 0.1 | 0.97 | 30 |
| 74 | 0.9 | 0.2 | 0.97 | 30 |
| 75 | 0.9 | 0.3 | 0.97 | 30 |
| 76 | 0.9 | 0.4 | 0.97 | 30 |
| 77 | 0.9 | 0.5 | 0.97 | 30 |
| 78 | 0.9 | 0.6 | 1 | 31 |
| 79 | 0.9 | 0.7 | 1 | 31 |
| 80 | 0.9 | 0.8 | 1 | 31 |
| 81 | 0.9 | 0.9 | 1 | 31 |